\documentclass[10pt,conference]{IEEEtran}
\IEEEoverridecommandlockouts
\usepackage{cite}
\usepackage{subcaption} 
\usepackage{graphicx}
\usepackage{amsmath,amssymb,amsfonts}
\usepackage{algorithmic}
\usepackage{graphicx}
\usepackage{textcomp}
\usepackage{xcolor}
\usepackage{hyperref}
\def\BibTeX{{\rm B\kern-.05em{\sc i\kern-.025em b}\kern-.08em
    T\kern-.1667em\lower.7ex\hbox{E}\kern-.125emX}}
\usepackage{listings}
\usepackage{xcolor} 

\begin{document}

\title{Fuzzy Accuracy Compensates for Label Subjectivity in Classification of Skin Tone Using Wearable Photoplethysmography Signals
\thanks{The project (22HLT01 QUMPHY) has received funding from the European Partnership on Metrology, co-financed from the European Union’s Horizon Europe Research and Innovation Programme and by the Participating States. Funding for NPL was provided by Innovate UK under the Horizon Europe Guarantee Extension, grant number 10084125.

\copyright~2026 IEEE.  Personal use of this material is permitted.  Permission from IEEE must be obtained for all other uses, in any current or future media, including reprinting/republishing this material for advertising or promotional purposes, creating new collective works, for resale or redistribution to servers or lists, or reuse of any copyrighted component of this work in other works.
}}

\author{\IEEEauthorblockN{Padmini Krishnadas}
\IEEEauthorblockA{\textit{Data Science and AI Department} \\
\textit{National Physical Laboratory}\\
Teddington, UK\\
Padmini.Krishnadas@npl.co.uk}
\and
\IEEEauthorblockN{Urs Hackstein}
\IEEEauthorblockA{\textit{Faculty of Life Science Engineering} \\
\textit{Mittelhessen University of Applied Sciences}\\
Giessen, Germany \\
Urs.Hackstein@lse.thm.de}
\and
\IEEEauthorblockN{Alen Bo\u{s}njakovi\'c}
\IEEEauthorblockA{\textit{Laboratory for Mass and Related Quantities} \\
\textit{Institute of Metrology of Bosnia and Herzegovina}\\
Sarajevo, Bosnia and Herzegovina \\
Alen.Bosnjakovic@met.gov.ba}
\and
\IEEEauthorblockN{Philip J. Aston}
\IEEEauthorblockA{\textit{Data Science and AI Department} \\
\textit{National Physical Laboratory}\\
Teddington, UK \\
Philip.Aston@npl.co.uk}}

\maketitle

\begin{abstract}
We consider the problem of classification of skin tone using photoplethysmography (PPG) signals with labels of the ordinal six-class Fitzpatrick skin tones. A typical accuracy for this task is a poor 40-55 \%. However, the labels are subjectively determined by comparing the skin with a colour chart, and hence contain widespread small-scale inaccuracies. By working with a ``fuzzy accuracy'', which deems a prediction of skin tone class to be correct if its difference from the labelled class is not greater than one, much higher accuracy is obtained which provides more convincing evidence that skin tone can be accurately predicted from PPG signals. Three machine learning approaches were used, namely deep learning or tree-based approaches on raw PPG signals, deep learning on image representations of the signals generated by the Symmetric Projection Attractor Reconstruction (SPAR) method, and machine learning on features extracted from the signals. The first method also employed a fuzzy version of the cross entropy loss function, which gave the best results. Tree-based models on raw signals give accuracies up to 55 \% and higher fuzzy accuracies up to 96 \%, while deep learning models on the SPAR images obtained lower results of 44 \% accuracy and 85~\% fuzzy accuracy. The machine learning on PPG features gave similar results to the SPAR method with accuracy of 42 \% and fuzzy accuracy of 87 \%. We have shown that classification of skin tone using PPG signals is possible with high fuzzy accuracy which implies that our modelling approach enables accurate prediction of skin tone class within at most one class of the observer's choice of class, from which we conclude that PPG signals are affected by skin tone in a discernible way.
\end{abstract}

\begin{IEEEkeywords}
fuzzy accuracy, fuzzy cross entropy, Fitzpatrick skin tone, photoplethysmography signals
\end{IEEEkeywords}

\section{Introduction}

Machine learning classification models are trained on data consisting of features and class labels, and the training process involves learning the mapping from features to class labels. This process assumes that the class labels are accurate and correct. However, in many medical disease diagnosis problems, the labels indicate the presence or absence of particular diseases and are determined by clinicians based on data from diagnostic tests (such as the electrocardiogram (ECG) for cardiovascular diseases) but clinicians do not always agree on the diagnosis as subjective judgement is required. Thus, there is the potential for a non-negligible proportion of records to be misclassified.

The problem we consider has much greater potential for label errors and consists of classification of skin tone from photoplethysmography (PPG) signals. PPG signals are generated by both pulse oximeters, which are widely used in a clinical environment, and by consumer wearable devices such as smart watches and rings. Both devices offer non-invasive continuous health monitoring, often deployed among vulnerable populations in intensive care or with pre-existing medical conditions, highlighting the need for accurate measurements. However, it was found during the Covid-19 pandemic that the blood oxygenation readings from pulse oximeters were overestimated for patients with darker skin tones which has the potential for incorrect decisions being made regarding treatment in these cases. An
independent report Equity in Medical Devices: Independent Review \cite{report} was commissioned in the UK after the pandemic which found "extensive evidence of poorer performance of pulse oximeters for patients with darker skin tones". PPG signals are generated by shining light onto the skin and measuring the light that is reflected back and so it is not surprising that the reflected light, and hence the PPG signal, is different for light and dark skin tones. We consider the classification of skin tone from PPG signals to investigate this dependence further using machine learning (ML). ML models have been shown to differentiate ethnicity from medical data that does not contain this information directly, making them well suited to the skin tone classification problem \cite{noseworthy2020assessing, kim2018riddle, gichoya2022ai}. If skin tone cannot be accurately classified using PPG signals by machine learning, it suggests that the signals do not depend on skin tone. Conversely, if machine learning does give an accurate classification of skin tone using PPG signals, then the signals are clearly dependent on skin tone in some way.

Skin tone is commonly classified according to the six-class Fitzpatrick skin tone scale \cite{Fitzpatrick1988} where the lightest skin tone is class I and the darkest is class VI. The Fitzpatrick scale was originally developed to classify people with white skin in order to select the correct UV dose for treating skin disease, not for characterising skin tone\cite{Fitzpatrick1988}. However, its use as a scale for skin tone is now ``nearly ubiquitous'' \cite{Monk2023} and there are a few PPG datasets that record the Fitzpatrick skin tone for subjects, along with other metadata, including the AuroraBP dataset \cite{AuroraBP} that we use. Another complication with this scale is that the classes, and their boundaries, are subjective and lack a formal, measurable definition. Moreover, for accurate classification, the skin tone should be measured using a spectrophotometer (or similar device) at the site of data collection (e.g.\ the wrist for a smart watch). However, this is not usual practice. Instead, skin tone is commonly assessed subjectively by comparing the skin with a colour chart showing the typical tone in each class. Using this comparison with a colour chart, there can often be uncertainty as to whether the particular skin under consideration should be in class $n$ or $n+1$ for some some $n$. Indeed, in an assessment of manual grading, there were many one class differences between the classes designated by the graders for a particular facial image, and differences of two or more classes also occurred \cite{Krishnapriya2022}. Such subjective assessment can result in a large number of misclassifications which correspond to inaccurate labels for our problem of skin tone classification.

A problem such as this with potentially widespread, small-scale inaccuracies in the labels can result in poor absolute accuracy when using a machine learning model trained on such data. As the labels could be described as ``fuzzy'' or uncertain, and are also ordinal, we define a ``fuzzy accuracy'', whereby a class prediction of $n$ is assumed to be correct if the assigned label is within one class of the prediction, i.e.\ $n-1$, $n$ or $n+1$. This fuzzy accuracy is clearly greater than or equal to the accuracy by definition, so of interest will be the magnitude of the increase in the fuzzy accuracy.

In related work, the influence of skin tone on machine learning–based predictions has recently been investigated in \cite{Aston2024}, where classification of high blood pressure from PPG signals was examined. This study showed that when a model was trained exclusively on records from individuals with Fitzpatrick skin tone class I (the lightest skin tone) and subsequently evaluated on data from the other skin tone classes, classification accuracy generally declined markedly for those classes in comparison with models which had been trained on data which included all skin tone classes. Despite this, many publicly available PPG datasets either do not record skin tone information or are heavily skewed towards the lighter skin tones, with relatively few samples from darker skin tones. This imbalance is present, for example, in the Aurora-BP dataset used in this work \cite{AuroraBP}. As a result, models trained primarily on data from lighter-skinned subjects may not generalise well to individuals with darker skin tones, raising concerns about potential performance disparities.

The objective of this study is therefore to investigate whether skin tone itself can be predicted from PPG signals using machine learning and evaluated with our concept of fuzzy accuracy, framed as a six-class classification problem based on the Fitzpatrick skin tone scale in order to provide evidence of a machine learning model's sensitivity to skin tone in PPG signals. Machine learning methods are well suited to identifying complex relationships between signals and target labels. Consequently, strong predictive performance on this task would suggest that PPG signals exhibit systematic differences across skin tone categories in which case PPG datasets used for other machine learning tasks should include a uniform distribution of skin tones to ensure that the results are not biased. Conversely, poor classification performance would indicate that machine learning models are unable to reliably distinguish between signals associated with different skin tones, implying that skin tone may have a limited effect on the PPG signals themselves in which case skin tone would not have to be considered when constructing datasets for other machine learning tasks involving PPG signals.

The code for this work is available at \\
\href{https://gitlab.com/qumphy/fuzzy_accuracy_ppg_skin_tone}{https://gitlab.com/qumphy/fuzzy\_accuracy\_ppg\_skin\_tone}.

\subsection{Data}

For this study, we used the AuroraBP dataset \cite{AuroraBP} which includes PPG waveforms (sampling frequency 500 Hz) collected using a wrist-worn optical device and Fitzpatrick skin tone labels for 823 participants (51.6~\% female). Two distinct protocols were used, namely auscultatory and oscillometric, and we combined both of these into a single dataset giving a combined total of 19,017 records. The signals were of varying length but were generally around 15 s long.

\begin{figure}[t!]
\centerline{\includegraphics[width=9cm]{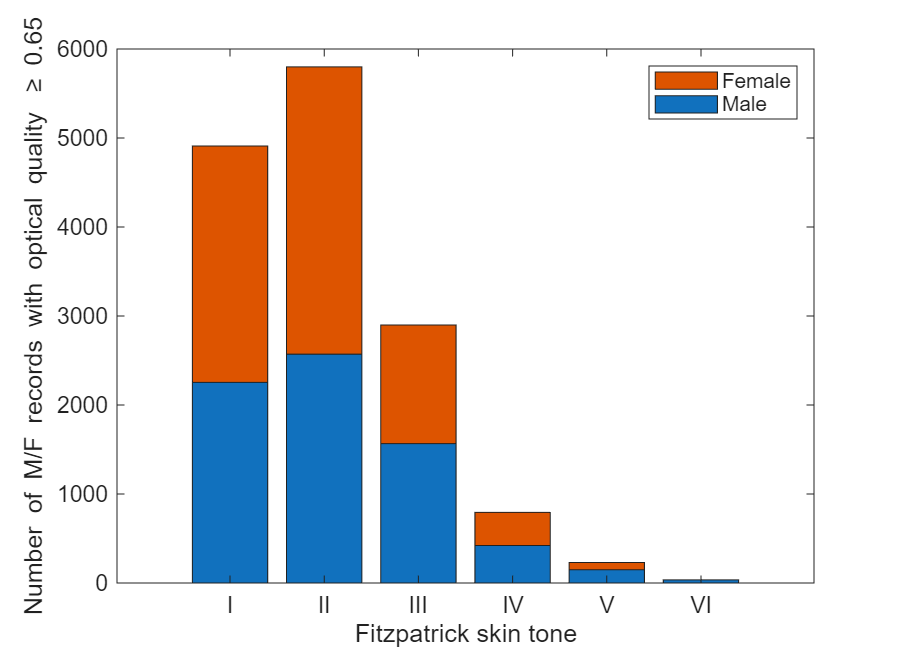}}
\caption{\footnotesize\textsc{The number of records in each skin tone class for which optical quality $\geq 0.65$.}}
\label{fig:aurora-optical-quality}
\end{figure}

The signals were of varying quality and a measure of ``optical quality" is included in the dataset. This measure takes into account artifacts, signal to noise ratio and consistency between pulses \cite{AuroraBP}. Records with optical quality $\geq 0.65$ were used in \cite{AuroraBP} so we used the same threshold. There are a total of 14,667 records satisfying this condition of which only 35 records are skin tone class VI.

The distribution of records with optical quality $\geq 0.65$ by skin tone class is shown in Fig.\ \ref{fig:aurora-optical-quality}, which shows that the dataset is highly imbalanced with the majority of subjects having lighter skin tones (92.8~\% in classes I, II and III). Also, there are no female subjects with skin tone class VI.

\section{Methods}

Three different machine learning approaches will be used for this classification problem in order to consider different input representations, namely deep learning using one-dimensional convolutional neural networks (CNN) or tree-based machine learning models taking the raw PPG signals as input, transformation of the PPG signals into Symmetric Projection Attractor Reconstruction (SPAR) attractor images \cite{Aston2018} which are used as input to a two-dimensional CNN, and machine learning using features extracted from the PPG signals. For all three methods, the normal accuracy will be reported together with our fuzzy accuracy, which we now define.

\subsection{Fuzzy Accuracy for Ordinal Fitzpatrick Labels}\label{fuzzyacc}

Self-reported Fitzpatrick skin tone is known to exhibit subjectivity and inter-annotator variability, with frequent near-adjacent label discrepancies \cite{tian2024shades, weir2025evaluating}. To compensate for this uncertainty when evaluating model performance, we evaluate our models using \emph{fuzzy accuracy}, a tolerance-aware metric tailored to ordinal classes.

Let $y_i \in \{0,\ldots,5\}$ denote the labelled class (I--VI mapped to 0--5) and $\hat{y}_i$ the predicted class for sample $i$. We define a tolerance radius $\tau \in \mathbb{N}$ and a weight vector $\mathbf{w} = (w_0,w_1,\ldots,w_\tau)$ with $1 = w_0 \ge w_1 \ge \cdots \ge w_\tau \ge 0$. The per-sample fuzzy match is
\[
m_i(\hat{y}_i,y_i;\tau,\mathbf{w}) \;=\; 
\begin{cases}
w_d, & d = |\hat{y}_i - y_i| \le \tau,\\
0,   & \text{otherwise}.
\end{cases}
\]
The \emph{fuzzy accuracy} is the mean fuzzy match over $N$ samples:
\[
\operatorname{Acc}_{\text{fuzzy}}(\tau,\mathbf{w}) \;=\; \frac{1}{N}\sum_{i=1}^{N} m_i(\hat{y}_i,y_i;\tau,\mathbf{w}).
\]
In this work, we report the special case $\tau=1$ with $(w_0,w_1)=(1,1)$, referred to as \emph{tolerant accuracy} ($\pm 1$), which treats adjacent-class predictions as correct. We also provide \emph{strict accuracy} ($\tau=0$, $w_0=1$) for reference.

In addition to considering fuzzy accuracy, we also included a \emph{fuzzy cross entropy} (neighbour label smoothing) computed against soft targets that allocate a small mass $\alpha$ to the immediate neighbours ($\pm 1$) of $y_i$, with edge classes having a single neighbour. For logits $\mathbf{z}_i\in\mathbb{R}^{6}$ and soft target $\mathbf{s}_i\in[0,1]^6$,
\[
\mathcal{L}_{\text{fuzzyCE}} \;=\; 
-\frac{1}{N}\sum_{i=1}^{N}\sum_{c=0}^{5} s_{i,c}\,\log\big(\operatorname{softmax}(\mathbf{z}_i)_c\big),
\]
where $s_{i,y_i}=1-\alpha_{\text{left}}-\alpha_{\text{right}}$, $s_{i,y_i-1}=\alpha_{\text{left}}$ (if $y_i>0$), and $s_{i,y_i+1}=\alpha_{\text{right}}$ (if $y_i<5$).
We employ $\mathcal{L}_{\text{fuzzyCE}}$ as a reported metric while optimising standard cross entropy for training. This design keeps training comparable across baselines yet quantifies robustness to plausible adjacent-class ambiguity. In addition to this, we report the Earth Mover's Distance (EMD) loss, which uses predicted probabilities of all classes and penalises missed predictions in accordance to a ground reference matrix quantifying class similarity \cite{hou2016squared, zhang2018emd}, making EMD well aligned with the underlying structure of skin tone gradations in comparison to a standard cross entropy loss.
This formulation is consistent with fuzzy evaluation ideas in the soft classification literature \cite{gomez2008determining,binaghi1999fuzzy,silvan2008sub} and recent work on ordinal-aware soft targets and calibration \cite{kim2024calibration, kang2023fuzzy}.

\subsection{Deep/Machine Learning Using Raw Signals}

For the first approach, using the raw PPG signals as input to a deep learning model (1d CNN), we first truncated all signals to the first 5,877 samples (11.75 s) to achieve uniformity as signal lengths vary, and the minimum length is 5,877 samples. 

Next, the signals were processed in two ways:
\begin{itemize}
    \item[i)] Deep learning models were trained using the raw signals.
    \item[ii)] The signals were filtered using a fourth order Chebyshev II bandpass filter (0.5–15 Hz cutoff, 20 dB stopband attenuation) \cite{Liang2018}, and then the same models were trained.
\end{itemize}

The following ML models were evaluated to compare performance:
\begin{itemize}
    \item Random forest \cite{pedregosa2011scikit}
    \item Gradient boosted decision trees \cite{GBBSP23}
    \item LeNet-1D \cite{D1}
    \item ResNet-1D50 \cite{D1}
    \item ResNet-1D101 \cite{D1}
    \item VGG16 \cite{D1}
\end{itemize}

A stratified 5-fold cross-validation by subject was employed to ensure that no signals from a subject were in both the training and test datasets. For Gradient boosted decision trees, multiple loss functions were tested, including the multinomial log-likelihood, a cross entropy loss function and two self-defined variants of multinomial log-likelihood and cross entropy respectively reflecting fuzzy accuracy. This was done in a similar way to the explanation given in Section \ref{fuzzyacc} using label smoothing and an indicator matrix which treats adjacent-class predictions as correct.

This is illustrated by the following definition of fuzzy cross entropy loss: 

\begin{lstlisting}
def fuzzy_loss(
    labels: npty.NDArray[np.int32],
    predictions: npty.NDArray[np.float32],
    weights: npty.NDArray[np.float32],
) -> np.float32:
    dimension = np.max(labels) 
    sum_exp_pred = np.sum(np.exp(predictions), axis=1)
    
    # Indicator matrix for Fuzzy accuracy
    indicator_matrix = np.zeros((labels.size, dimension))
    for i, label in enumerate(labels-1):  
        if label > 0:
            indicator_matrix[i, label - 1] = 0.5  # x-1
        indicator_matrix[i, label] = 1  # x
        if label < dimension - 1:
            indicator_matrix[i, label + 1] = 0.5  # x+1

    label_exp_pred = np.exp(np.sum(predictions * indicator_matrix, axis=1))
    return -np.sum(np.log(label_exp_pred / sum_exp_pred)) / len(labels)
\end{lstlisting}

For Random forest and Gradient boosted decision trees, both standard accuracy and fuzzy accuracy are also computed for subsets restricted to male or female subjects, although the choice of classifier was motivated by their performance on the whole dataset.

\subsection{Deep Learning with SPAR Attractor Images}
\label{sec:DL with SPAR}
\subsubsection{SPAR Attractor Image Generation}

\begin{figure}[t]
\centerline{\includegraphics[width=9cm]{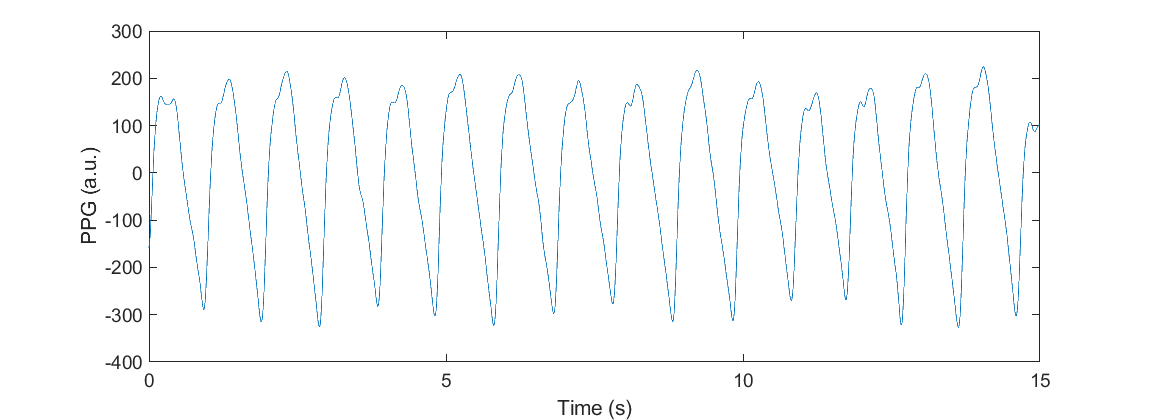}}
\centerline{\includegraphics[width=7cm]{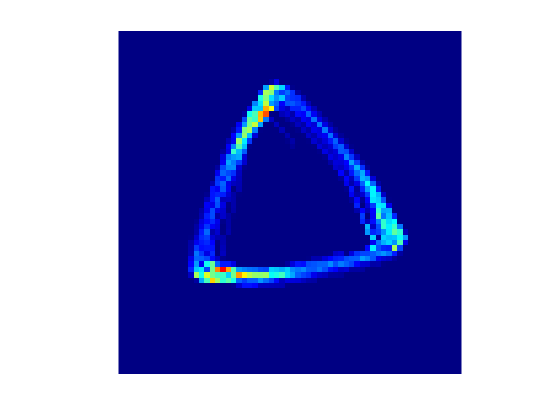}}
\caption{\footnotesize\textsc{A PPG signal (top) and the corresponding SPAR attractor (bottom).}}
\label{fig:SPAR}
\end{figure}

The Symmetric Projection Attractor Reconstruction (SPAR) method converts an approximately periodic signal into a compact image which encapsulates the morphology and variability of the signal \cite{Aston2018} (see Fig. \ref{fig:SPAR}). The SPAR attractor images can be used as input to a two dimensional CNN for classification of the Fitzpatrick skin tone.

For this approach, we used only the first 15 s of longer signals and all of the signals for shorter signals. To generate SPAR images, we first filtered the signals using a fourth order high pass Chebyshev II filter using a cut-off frequency of 0.5~Hz \cite{Liang2018}. This removes any baseline wander from the signals which can affect the SPAR attractors. The size of the attractor is determined by the amplitude of the signals \cite{Aston2018}. Thus, we scaled the signals in the vertical direction by the amplitude so that all signals had amplitude 1 resulting in consistently sized attractors. We then generated SPAR attractor images for each signal using a $64\times 64$ grid and three delay coordinates, resulting in an approximate threefold rotational symmetry in the attractors.

\subsubsection{Deep Learning}

For SPAR attractor images, skin tone classification is formulated as an image classification task. We compared the performance of ResNet50, ResNet152 and VGG-16 models for classifying Fitzpatrick skin tones. We used ImageNet-pretrained ResNet-50 ResNet152 and VGG-16 backbones. ResNet models employed bottleneck residuals with final feature dimensions of 2048, while VGG-16 uses sequential 3x3 convolutions followed by a 4096-unit classifier. For all architectures, we replaced the final classification layer with a task specific output head, mapping to the six skin tone classes and applied differential learning rates to the backbone and classifier, and an Adam optimiser.  Prior to training, data augmentation techniques were applied to SPAR images to increase dataset diversity, improve generalisation, and mitigate overfitting. Augmentations also enhance model robustness by effectively expanding the training set \cite{krishnadas2022, Mikołajczyk2018, elgendi2021, yang2023}. Augmentations considered are: random horizontal and vertical flips, rotations and resized crop. To address uncertainty in Fitzpatrick skin tone labels, we adopt fuzzy accuracy for multi-class evaluation, implemented with the Earth Mover's Distance (EMD) loss function. 

\subsection{Machine Learning Using PPG Features}

We used the PulseAnalyse code \cite{CharltonPulseAnalyse} to extract PPG features from the raw signals. Further details about these PPG features are provided in \cite{Charlton_2018}. This function is designed to analyse arterial pulse waves, including blood pressure and PPG signals.

We applied a fourth-order Chebyshev Type II band-pass filter (0.5--15 Hz) to the raw signals to remove low-frequency drift and high-frequency noise before passing them to the PulseAnalyse function. We extracted $66$ features, including $39$ fiducial point–based features, $15$ amplitude-based features, and $12$ cardiovascular indices, and organised them into a feature table with dimension $14{,}667 \times 66$. Raw signals with optical quality $\geq 0.65$ were used, and all missing values were replaced by zeros. Furthermore, we were not able to process all the raw signals using the PulseAnalyse function, with approximately 0.3 \% (41 signals) not being processed, and additional effort will be invested to understand this issue.

We applied three supervised machine learning methods to these features, namely neural network, classification trees, and support vector machine. The neural network involved two fully connected layers, the first having 20 nodes, followed by ReLU activations.

\section{Results}

\subsection{Deep/Machine Learning Using Raw Signals}

\begin{figure}[t]
    \centering
    \includegraphics[width=\linewidth]{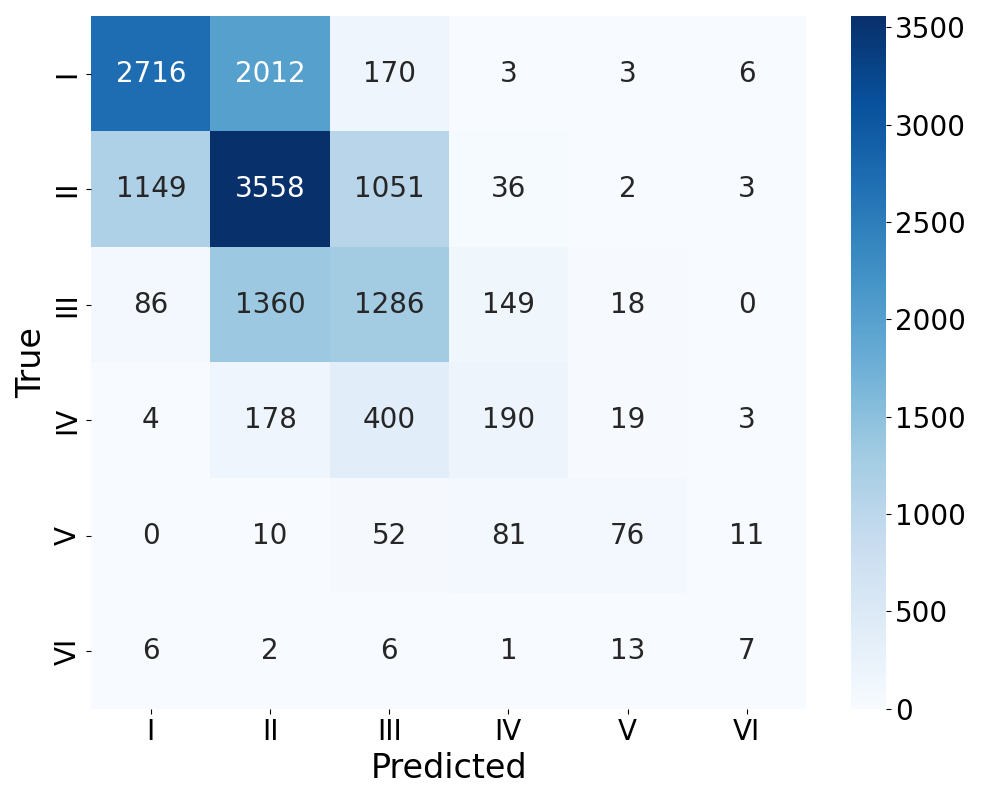}
    \caption{\footnotesize\textsc{Confusion matrix over 5-fold cross validation for the Gradient boosted decision tree model with fuzzy cross entropy loss function trained on raw signals.}}
    \label{rawcm}
\end{figure}

The values obtained for the accuracy and fuzzy accuracy for the different classifiers and loss functions when using a tree-based model on raw signals are shown in Table \ref{tableraw}. Similarly, Table \ref{tablefiltered} shows analogous results for the filtered signals. The other (CNN-based) classifiers (LeNet-1D, ResNet-1D50, ResNet-1D101 and VGG-16) showed significantly lower results for both accuracy and fuzzy accuracy. The evaluation of the tree-based models (i.e.\ Random forest and Gradient boosted decision trees) on female or male only subsets delivered similar values for accuracy and fuzzy accuracy with a difference that was always less than 0.04, which indicates that there is no gender bias in these results. We note that the fuzzy cross entropy loss function gave the best fuzzy accuracy for the raw signals, whereas the Random forest approach gave the best fuzzy accuracy for the filtered signals. However, the best accuracy was obtained using Gradient boosted decision trees with the multinomial log likelihood loss function for the raw signals or the Random forest approach for the filtered signals.\\
An exemplary confusion matrix combined over all folds for the case of Gradient boosted decision trees with fuzzy cross entropy loss function with 5-fold cross validation by subject is shown in Figure \ref{rawcm}. (Note that the numbers indicate classification by record not by subject.)

\begin{table}[!t]
\caption{\footnotesize\textsc{Accuracy and fuzzy accuracy values for different classifiers and loss functions applied to the raw signals. The best results are shown in bold.\\
GBT = Gradient boosted decision tree, RF = Random forest classifier}}
\label{tableraw}
\centering
\begin{tabular}{|l|l|c|c|}
\hline
&&& \textbf{Fuzzy}\\
\bfseries Classifier & \bfseries Loss Function & \bfseries Accuracy & \bfseries Accuracy \\ \hline
     GBT & Cross entropy & 0.5502 & 0.9561 \\ \hline
     GBT & Fuzzy cross entropy & 0.5336 & \textbf{0.9594} \\ \hline
     GBT & Multinomial log likelihood & \textbf{0.5503} & 0.9561 \\ \hline
     GBT & Fuzzy multinomial log likelihood & 0.4585 & 0.9339 \\\hline
     RF  & Standard & 0.5332 & 0.9531 \\ \hline
\end{tabular}
\vspace{0.5cm}
\caption{\footnotesize\textsc{Accuracy and fuzzy accuracy values for different classifiers and loss functions applied to the filtered signals. The best results are shown in bold.\\
GBT = Gradient boosted decision tree, RF = Random forest classifier}}
\label{tablefiltered}
\centering
    \begin{tabular}{|l|l|c|c|}
    \hline
    &&& \textbf{Fuzzy} \\
     \bfseries Classifier & \bfseries Loss Function & \bfseries Accuracy & \bfseries Accuracy \\ \hline
     GBT & Cross entropy & 0.5427 & 0.9563 \\ \hline
         GBT & Fuzzy cross entropy & 0.5369 & 0.9554 \\ \hline
     GBT & Multinomial log likelihood & 0.5427 & 0.9563 \\ \hline
     GBT & Fuzzy multinomial log likelihood & 0.4440 & 0.9264 \\\hline
     RF  & Standard & \textbf{0.5431} & \textbf{0.9564} \\ \hline
    \end{tabular}
\vspace{0.5cm}

\caption{\footnotesize\textsc{Overall performance metrics from ResNet50, ResNet152 and VGG-16 for multiclass skin tone classification using cross entropy loss and fuzzy accuracy. The best accuracy and fuzzy accuracy are shown in bold.}}
    \label{tab: SPARQ multiclass classification}
    \centering
    \begin{tabular}{|l|l|c|c|c|}
    \hline
    && \textbf{Loss}  &  & \textbf{Fuzzy} \\
     \textbf{Classifier} & \textbf{Loss function} & \textbf{Score} & \textbf{Accuracy} & \textbf{Accuracy}  \\ \hline
     ResNet50 & Cross entropy & 1.2138 & 0.4567 & 0.8435  \\ \hline
     ResNet152 & Cross entropy & 1.2071  & 0.4439 & 0.8450 \\ \hline
     VGG-16 & Cross entropy & 1.2111 & \textbf{0.4603} & 0.8459 \\ \hline
     ResNet50 & Earth Mover Distance & 0.0761 & 0.4673 & 0.8498 \\ \hline
     ResNet152 & Earth Mover Distance & 0.0783 & 0.4339 & 0.8504 \\\hline
     VGG-16 & Earth Mover Distance &0.076  & 0.4621 & \textbf{0.8588} \\ \hline
    \end{tabular}
\vspace{0.5cm}

\caption{\footnotesize\textsc{Accuracy and fuzzy accuracy values for different classifiers and loss functions applied to the PPG features. The best results are shown in bold.}}
\label{tableFeature}
\centering
\begin{tabular}{|l|l|c|c|}
\hline
&&& \textbf{Fuzzy}\\
\bfseries Classifier & \bfseries Loss Function & \bfseries Accuracy & \bfseries Accuracy \\ \hline
     Trees & Classification error & 0.3473 & 0.8100 \\ \hline
     Neural Network & Cross entropy & \textbf{0.4219} & \textbf{0.8705} \\ \hline
     Support Vector Machine & Classification error & \ 0.4024 & 0.8638\\ \hline
\end{tabular}
\end{table}

\subsection{Deep Learning using SPAR images}
\begin{figure}[h!]
    \centering
    \includegraphics[width=1.00\linewidth]{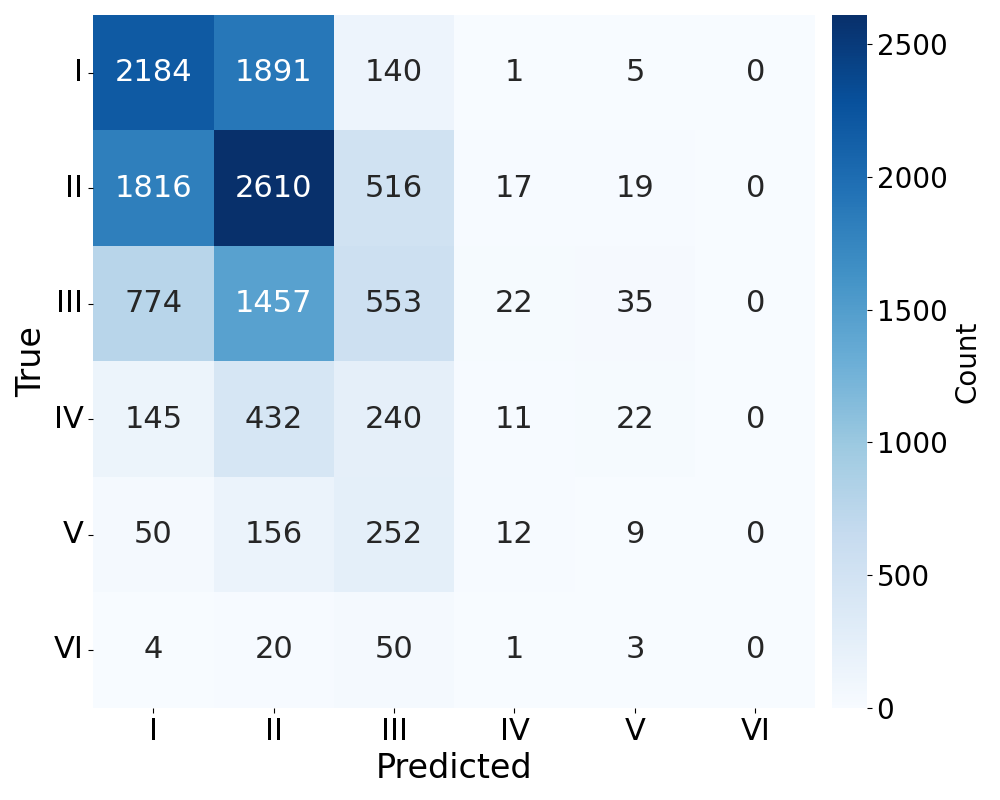}
    \caption{\footnotesize\textsc{Confusion matrix over 5-fold cross-validation using VGG16 with EMD loss trained on SPAR images.}}
    \label{fig:VGG16 spar conf mat}
\end{figure}

We assessed the ResNet50, ResNet152 and VGG-16 models using fuzzy accuracies, implemented with 5-fold cross-validation. All models reported similar classification accuracies, with VGG-16 giving a marginally higher fuzzy accuracy of 0.8588. Table \ref{tab: SPARQ multiclass classification} lists the performance of all models when classifying the Fitzpatrick skin tones. Figure \ref{fig:VGG16 spar conf mat} shows the confusion matrix across all 5 folds when using the VGG-16 model. This illustrates the model's capability to discriminate between classes I, II and even class III to some extent, with many misclassifications. However, the model was unable to identify records for classes IV, V and VI with any accuracy. This is likely due to a combination of imbalance in the dataset causing the model to favour learning light skin tones over dark as well as very small sample sizes effectively rendering classes V and VI as statistical outliers. 

\subsection{Machine Learning Using PPG Features}

\begin{figure}[h!]
    \centering
    \includegraphics[width=1.00\linewidth]{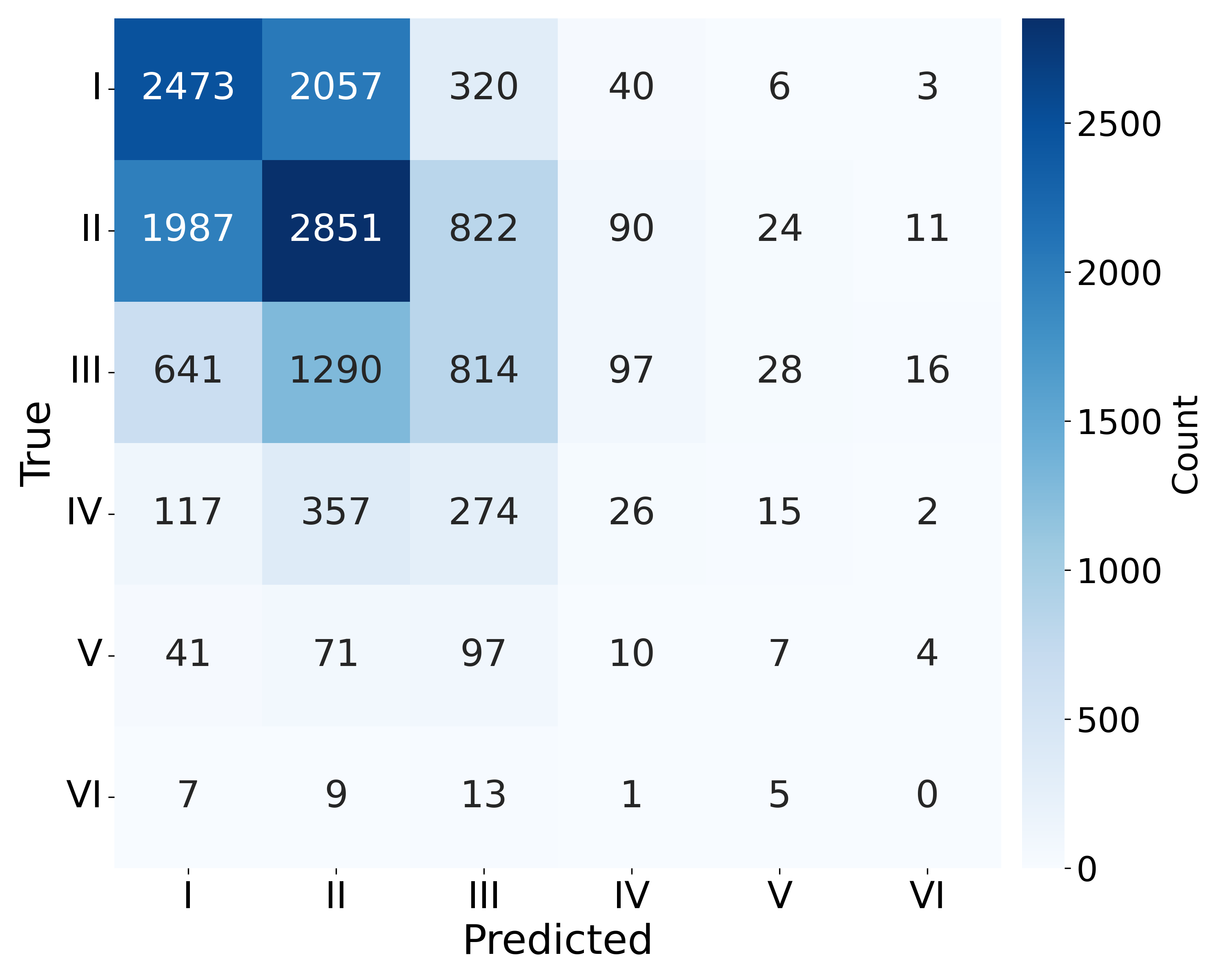}
    \caption{\footnotesize\textsc{Confusion matrix over 5-fold cross-validation for the neural network model using PPG Features.}}
    \label{fig: NN conf mat}
\end{figure}

Neural Networks and SVM methods achieved almost the same accuracy and the same fuzzy accuracy on PPG features (see Table \ref{tableFeature}) while Trees reported lowest performance metrics. The highest fuzzy accuracy was obtained using the neural network method. The results are presented using a confusion matrix that combines all folds for the neural network method with five-fold cross-validation performed by subject, as shown in Figure \ref{fig: NN conf mat}. The numbers represent classification by record rather than by subject.

\section*{Conclusions}

We have shown that in order to deal with widespread small-scale inaccuracies in the skin tone labels for our classification problem, the introduction of a fuzzy accuracy provides a metric that scores much higher than the normal accuracy, thus showing that machine learning prediction of skin tone class can be performed very accurately within one class of the subjective class label. Moreover, the use of a fuzzy cross entropy also resulted in a (marginally!) higher fuzzy accuracy.

The best results for this classification problem when considering fuzzy accuracy were obtained by tree-based models using raw signals as input. In particular, the best fuzzy accuracy was obtained using a Gradient boosted decision tree model with a fuzzy cross entropy loss when working with the raw signals, or a Random forest model when working with filtered raw signals. There was very little difference between the results using raw and filtered signals and predictions of the under represented darker skin tones was reasonable. The SPAR and PPG features methods gave fuzzy accuracies approximately 10 \% lower, and with poor prediction of the darker skin tones, with the dominant lighter skin tone classes being favoured. However, we note that the SPAR method also normalised the amplitudes of the signals in order to give comparably sized attractors and so any variation in amplitude with skin tone is lost in this case. Since the PPG signals are generated by light passing through the skin, it is very likely that the amplitude of the signals will be skin tone dependent.

The classification results that we have presented support our concept of fuzzy accuracy. The values for the accuracy are significantly higher than the uniform value of $16.7 \%$ which would arise from prediction of the classes if there was no dependence of the PPG signals on skin tone, but they are not high enough to give accurate prediction of the labelled Fitzpatrick skin tone of the subject. However, the fuzzy accuracies are reaching up to almost 96 \%, strongly supporting the hypothesis that skin tone classes can be determined from PPG signals up to a deviation of $\pm 1$. This also suggests that if the skin tone was measured more precisely, resulting in more accurate labels, then the prediction accuracy would be much better also. 

The Aurora BP dataset that we used for this study is highly imbalanced with regard to skin tone, with very few records with the darker skin tones. Future work could include repeating this study with a more balanced dataset.

A consequence of this work is that skin tone should be taken into consideration when constructing PPG datasets for machine learning, and in the calibration and use of wearable devices and pulse oximeters. Failure to do so could result in biased results and, potentially, consequent harm to the wearer.

\bibliographystyle{IEEEtran}
\bibliography{references}
\end{document}